\documentclass[letterpaper, 10 pt, conference]{ieeeconf}  %

\IEEEoverridecommandlockouts                              %
\usepackage{amsmath,amssymb,amsfonts}
\usepackage{bm}
\usepackage{algorithm2e}
\usepackage{algpseudocode}
\usepackage{graphicx}
\usepackage{svg}
\usepackage{caption}
\usepackage{subcaption}
\DeclareMathOperator*{\argmin}{arg\,min}

\usepackage{textcomp}
\usepackage{xcolor}
\def\BibTeX{{\rm B\kern-.05em{\sc i\kern-.025em b}\kern-.08em
    T\kern-.1667em\lower.7ex\hbox{E}\kern-.125emX}}
\SetKw{KwRequire}{Require:}

\title{\LARGE \bf
Tactile-based Active Inference for \\Force-Controlled Peg-in-Hole Insertions
}

\author{Tatsuya Kamijo$^{1}$, 
        Ixchel G. Ramirez-Alpizar$^{2}$, 
        Enrique Coronado$^{2}$, 
        Gentiane Venture$^{1,3}$%
\thanks{$^{1}$Department of Mechanical Engineering, Faculty of Engineering, The University of Tokyo, Japan
        {\tt tatsukamijo@g.ecc.u-tokyo.ac.jp}}%
\thanks{$^{2}$Automation Research Team, Industrial CPS Research Center, National Institute of Advanced Industrial Science and Technology (AIST), Japan%
        }%
\thanks{$^{3}$CNRS-AIST JRL (Joint Robotics Laboratory), National Institute of Advanced Industrial Science and Technology (AIST), Japan%
        }%
}

\begin{document}
\maketitle

\begin{abstract}
Reinforcement Learning (RL) has shown great promise for efficiently learning force control policies in peg-in-hole tasks. However, robots often face difficulties due to visual occlusions by the gripper and uncertainties in the initial grasping pose of the peg. 
These challenges often restrict force-controlled insertion policies to situations where the peg is rigidly fixed to the end-effector.
While vision-based tactile sensors offer rich tactile feedback that could potentially address these issues, utilizing them to learn effective tactile policies is both computationally intensive and difficult to generalize. 
In this paper, we propose a robust tactile insertion policy that can align the tilted peg with the hole using active inference, without the need for extensive training on large datasets. Our approach employs a dual-policy architecture: one policy focuses on insertion, integrating force control and RL to guide the object into the hole, while the other policy performs active inference based on tactile feedback to align the tilted peg with the hole.
In real-world experiments, our dual-policy architecture achieved 90\% success rate into a hole with a clearance of less than 0.1 mm, significantly outperforming previous methods that lack tactile sensory feedback (5\%).
To assess the generalizability of our alignment policy, we conducted experiments with five different pegs, demonstrating its effective adaptation to multiple objects.
\end{abstract}

\section{Introduction}
Broadening the application of industrial robots necessitates the ability to safely execute precise contact-rich manipulation tasks, such as peg insertion. Yet, the achievement of safe and robust execution of insertion tasks is challenging due to grasp uncertainty, visual occlusions by the gripper, and complex physical interaction between the grasped object and the environment.

Early work on peg-in-hole tasks leverage mathematical or geometrical model of the environment \cite{pegonhole, assemstrateg89}. 
In recent years, learning-based approaches have shown great success in various manipulation tasks by learning complex behaviours, thus avoiding the need to model physical interactions between the manipulated object and the environment \cite{review_rl_contact23}. Reinforcement Learning (RL) methods have especially been shown to be effective in object insertion tasks. 
While RL policies can successfully learn peg-in-hole insertion tasks, most of the work either consider the peg to be part of the robot's end-effector or fix it to the gripper, assuming no in-hand slippage occurs during the insertion process \cite{learning_forcecontrol20}.
This paper addresses this assumption by using feedback signal from the vision-based tactile sensors. 

\begin{figure}[tbp]
    \centering
    \includegraphics[width=\linewidth]{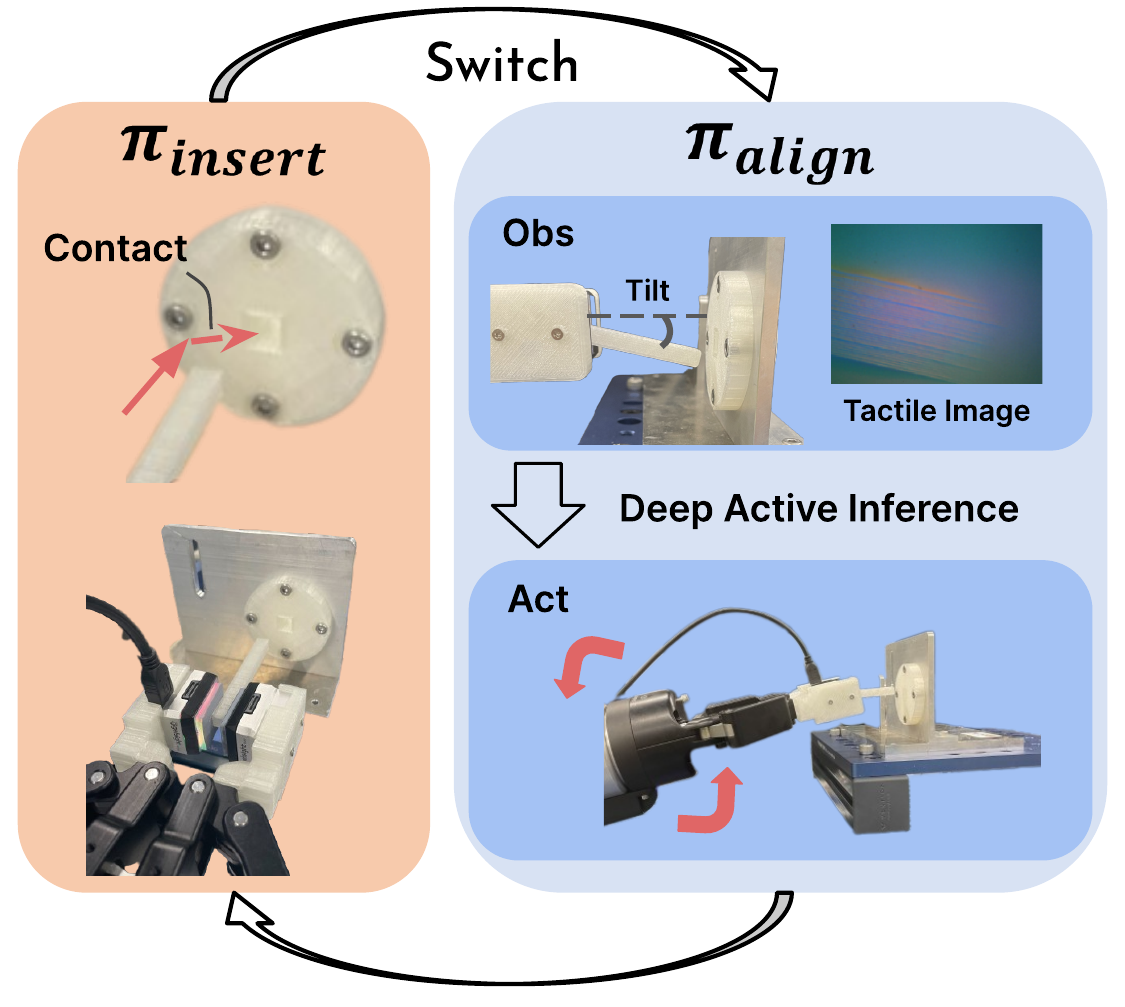}
    \caption{Proposed dual-policy system for robust insertion. It leverages deep active inference for tactile pose alignment.}
    \label{fig:dual}
\end{figure}

We propose a safe tactile insertion policy that leverages vision-based tactile sensors for robot pose alignment. Our method combines the strengths of force control and active inference, a promising neuroscience theory that unifies perception and action under a single objective of minimizing free energy. 
In our system, the agent predicts tactile sensation from its own internal state, namely the tilt of the peg. This internal state (i.e. inferred tilt) is updated in a way that minimizes discrepancies between predicted and current tactile image. The key insight is that by formulating tactile perception under free energy minimization problem, the active inference agent can adapt to diverse objects without extensive training on vast datasets.
Our approach consists of two policies as depicted in Fig. \ref{fig:dual}: 
\begin{itemize}
    \item \textbf{Insertion policy} that executes force-controlled policy learned by RL in simulation to insert the grasped peg.
    \item \textbf{Alignment policy} that employs active inference based on tactile feedback to change the robot pose.
\end{itemize}

The insertion policy handles the main process of inserting the peg. When the peg slips while in contact with the environment, the alignment policy takes over. It executes active inference to adjust the robot's end effector pose in a way that aligns the peg with the hole.

This paper presents three main contributions. 
First, we propose a novel deep active inference approach to tactile pose alignment. Building on the free energy principle, this approach adapts to various objects without a pre-trained model built on a large dataset or prior knowledge of their shape. 
Second, we introduce the self-data augmentation method to realize deep active inference without collecting extensive data in the real world.
Third, we combine a force-controlled insertion policy and the active inference-based tactile pose alignment policy. This dual-policy system is experimentally validated with the UR5 e-series robotic arm and the Gelsight Mini \cite{gelsight17} tactile sensor.

\section{Background and Related Work}
\subsection{Learning Contact-Rich Manipulation Tasks}
Learning force control by RL has shown significant promise for handling complex, contact-rich manipulation tasks \cite{review_rl_contact23}. Beltran-Hernandez et al. \cite{cristian20} proposes a learning-based force control framework that combines RL techniques with traditional force control methods. The authors use transfer-learning techniques (Sim2Real) and domain randomization to close the reality gap.
Although these methods have advanced work in object-insertion tasks, they typically assume that the peg is rigidly fixed to the robot's end-effector to prevent in-hand slippage during the insertion process. However, visual occlusions by the gripper often make it difficult for a robot to receive feedback about the physical state around the end-effector. 

Recent advances in high-resolution vision-based tactile sensors \cite{gelsight17, gelslim18, digit20, omnitact20} have enabled robots to obtain rich tactile information directly from their grippers. 
Approaches leveraging tactile information for manipulation tasks include tactile mapping \cite{Bauz2019TactileMA}, pose estimation \cite{Bauz2corl, poseaamas22} and supervised learning \cite{dense19, safe23}.
Learning a tactile insertion policy through RL is one way to utilize this tactile information. Dong et al. \cite{tactileRL21} study different design choices for tactile-based feedback insertion policies and propose an RL-based insertion policy that incorporates curriculum training and tactile flow representation. They also argue that raw tactile RGB images contain detailed object-specific features, which can cause a learning-based agent to easily overfit to the training objects. While tactile flow representation is effective in dynamical situation where markers on the surface change position in response to actions, it is not the best option when you need to obtain the static state of the peg such as tilt.  
Contrary to this, our work demonstrates that static RGB images can also generalize across various objects when you use active inference in combination with contact area estimation. We utilize the neural network architecture proposed in \cite{canfnet23} to estimate the contact area. This estimated shape is then used as input for our active inference controller.

\subsection{Free Energy Principle and Active Inference}
\indent The free energy principle is a theoretical framework from neuroscience that proposes that living systems seek to minimize a statistical quantity known as free energy \cite{friston10}. Building upon this theory, body perception can be modeled as inferring the hidden state $\bm{z}$ from the sensory observation $\bm{o}$. 
 The objective in perception is to find a hidden state $\bm{z}$ that is consistent with Bayes' rule given an observation $\bm{o}$: \\
\begin{equation}
    p(\bm{z} | \bm{o}) = \frac{p(\bm{o} | \bm{z}) p(\bm{z})}{p(\bm{o})}.
\end{equation}
\noindent Here, $p(\bm{z} | \bm{o})$ is the posterior probability of the internal state $\bm{z}$ given a sensory observation $\bm{o}$. To avoid calculating the marginal likelihood $p(\bm{o}) = \int_z p(\bm{o} | \bm{z}) p(\bm{z})d\bm{z}$, which is intractable when the dimension of the hidden state is large, a reference distribution, also referred to as the recognition density, $q(\bm{z})$ is introduced. By minimizing the Kullback-Leibler divergence $D_{\mathrm{KL}}$ between the true posterior and the recognition density, the hidden state $\bm{z}$ can be optimized.
\begin{equation} \label{eq:Dkl}
    D_{\mathrm{KL}} \left( q(\bm{z}) \parallel p(\bm{z} | \bm{o}) \right) 
    = \int q(\bm{z}) \ln \dfrac{q(\bm{z})}{p(\bm{o}, \bm{z})}  d\bm{z} 
    = F + \ln p(\bm{o})
\end{equation}
Rearranging (\ref{eq:Dkl}), free energy can be written as
\begin{equation} \label{eq:F}
    F = D_{\mathrm{KL}} \left( q(\bm{z}) \parallel p(\bm{z} | \bm{o}) \right) - \ln p(\bm{o}).
\end{equation}
The first term is associated with perception, where the agent performs Bayesian inference to better model its environment. The second term $-\ln p(\bm{o})$ represents the sensory surprise and is related to the action component, where the agent acts to minimize this sensory surprise and thereby achieve desired sensory outcomes. \\
\indent Active inference is a theory that outlines the specific mechanism by which a system acts on its world to change sensory inputs, thereby minimizing free energy. From Equation (\ref{eq:F}), minimizing $D_\mathrm{KL}$ is equivalent to minimizing free energy $F$ when the observation $\bm{o}$ is fixed. This is known as perceptual inference. 
Perceptual inference can tightly constrain the level of surprise by approximating the world but cannot lessen the surprise within the observations themselves $-\ln p(\bm{o})$. 
In active inference, the agent can decrease this sensory surprise by acting upon the environment to change sensory observations $\bm{o}$, thereby minimizing free energy. 
Thus, both perception and action can be done under the single free energy minimization:
\begin{align}
    \bm{z} &= \argmin_{\bm{z}} F(\bm{z}, \bm{o}), \\
    \bm{a} &= \argmin_{\bm{a}} F(\bm{z}, \bm{o}(\bm{a})).
\end{align}
\indent Active inference has been recently applied in the field of robotics \cite{survey_lanillos2021}. 
In applying active inference to robotics, generative functions are critical. Generative functions are learned mappings that approximate the internal hidden state dynamics and how sensory observations are generated from this hidden state, essentially serving as the model's understanding of the underlying causal structure of its environment.
The work in \cite{PixelAI2020} presents a deep learning approach to learn the generative function, proposing PixelAI that deals with high-dimensional RGB input for body perception and action. This type of learning-based method called Deep Active Inference (Deep AIF) in general has extended the active inference framework to high-dimensional inputs by learning generative functions \cite{deepaif18, deepaif_MC20, deepaif_POMDP20, deepaif_vpg20}. As for the choice of hidden state, previous work that apply active inference to robotics often assume the robots' joint angles as internal state \cite{adaptivebody18, PixelAI20, novelmanip20, empiricalhumanoid22}. This makes sense as joint angles are the only factors that bring change to visual sensations in the context of robot body perception and action. Our work sets apart from these approaches in that we treat external objects within the framework of active inference. We adopted the tilt of the grasped object as the internal state and learned the generative model that predicts tactile sensations from this tilt. Further details about our approach are presented in Section \ref{methodology}.

\section{Problem Statement} \label{problem statement}
We tackle the peg-in-hole task with a 6 DoF UR5 e-series robot arm equipped with vision-based tactile sensors within its fingers. 
The objective is to safely insert various objects into corresponding holes given the initial and the goal end-effector pose. 
During the whole insertion process, the robot has access to its joint angles, force-torque readings and tactile readings from one of the tactile sensors. 
No visual information was given to the robot in this work.
In the experiments we made the following assumptions: \\
\begin{enumerate}
    \item The width of the peg to be grasped is small enough such that a ``tilt" can be perceived in the tactile image.
    \item The inferred tilt obtained from the iteration of free energy minimization is accurate enough to be used as an observation of the robot. 
    \item The target pose of the end effector is known.
\end{enumerate}

\section{Methodology} \label{methodology}
\subsection{Overview}
To achieve robust and adaptive insertion, we propose a dual-policy structure, as depicted in Fig. \ref{fig:dual}. The system initially employs a learned force control policy to insert the grasped peg into the hole while avoiding excessive force \cite{cristian20}.  
When the grasped peg slips and tilts within the gripper due to physical contact with the environment, the system transitions to executing active inference. In this phase, the agent infers the tilt of the peg and uses this inference to decide what pose to take.
 \begin{figure}
    \centering
    \includegraphics[width=\linewidth]{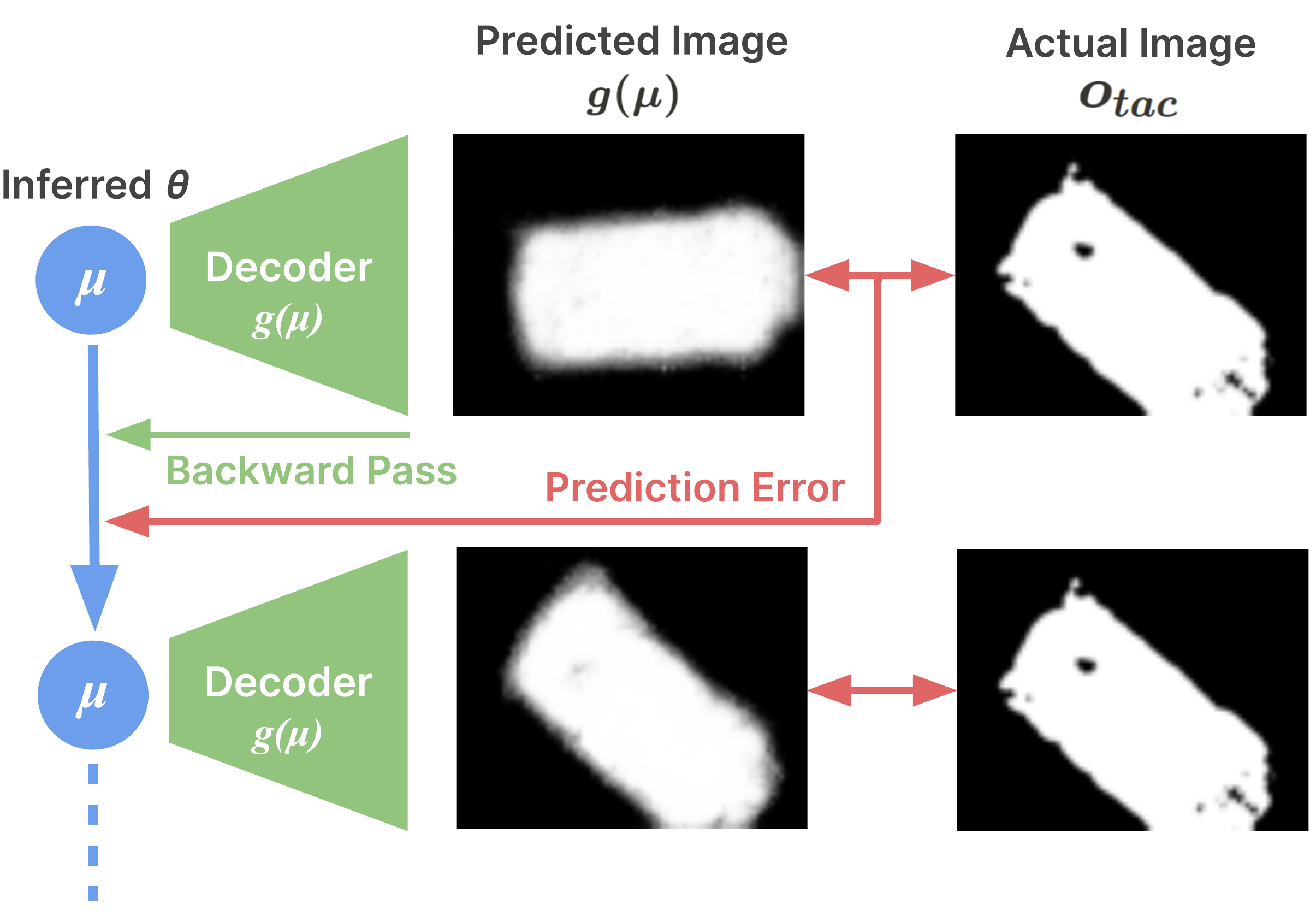}
    \caption{The internal state is updated in real-time to minimize free energy, thereby reducing discrepancies between the predicted and the actual tactile images of the contact area estimation. Note that vague contact area estimation is enough for performing $\mu$ updates based on prediction error minimization. The agent calculates the inverse kinematics based on the inferred tilt $\mu$ and adjusts the end-effector pose to align the peg with the hole. }
    \label{fig:architecture}
\end{figure}
\subsection{Insertion Policy}
Our force-controlled insertion policy is based on \cite{cristian20}. We use a PID-based parallel position-force control scheme with a selection matrix to determine the degree of control exerted on position and force along each direction. The control command given to the robot $\bm{x}_c$ is composed of a PD action on position, a PI action on force, a selection matrix $S$, and a position action from the RL policy $\bm{a}_x$:
\begin{equation}
    \bm{x}_c = S(K_p^x \bm{x}_e + K_d^x \bm{\dot{x}_e}) + \bm{a}_x + (I - S)(K^f_pF_e + K^f_i\int F_e dt),
\end{equation}
\noindent where $F_e$ and $\bm{x}_e$ represent the target force and position error, respectively. $K_p^x$, $K_d^x$, $K_p^f$, and $K_i^f$ are the controller gain parameters.
In this work, we exclude rotational action from the control command, as the alignment policy handles that aspect.
All the PID and selection matrix parameters are learned by an RL policy in simulation and transferred to the real world (Sim2Real). Domain randomization \cite{domain17} is used to close the reality gap between simulation physics and the real-world dynamics. For more details, see \cite{cristian20}.

\subsection{Alignment Policy}

Fig. \ref{fig:architecture} outlines the agent's perception of tactile observations based on its internal state. In the following sections, we detail our active inference model and describe how it is implemented for tactile sensory data.

\subsubsection{\textbf{Active Inference Model}} \label{aifmodel}

\begin{figure}
    \centering
    \includegraphics[width=\linewidth]{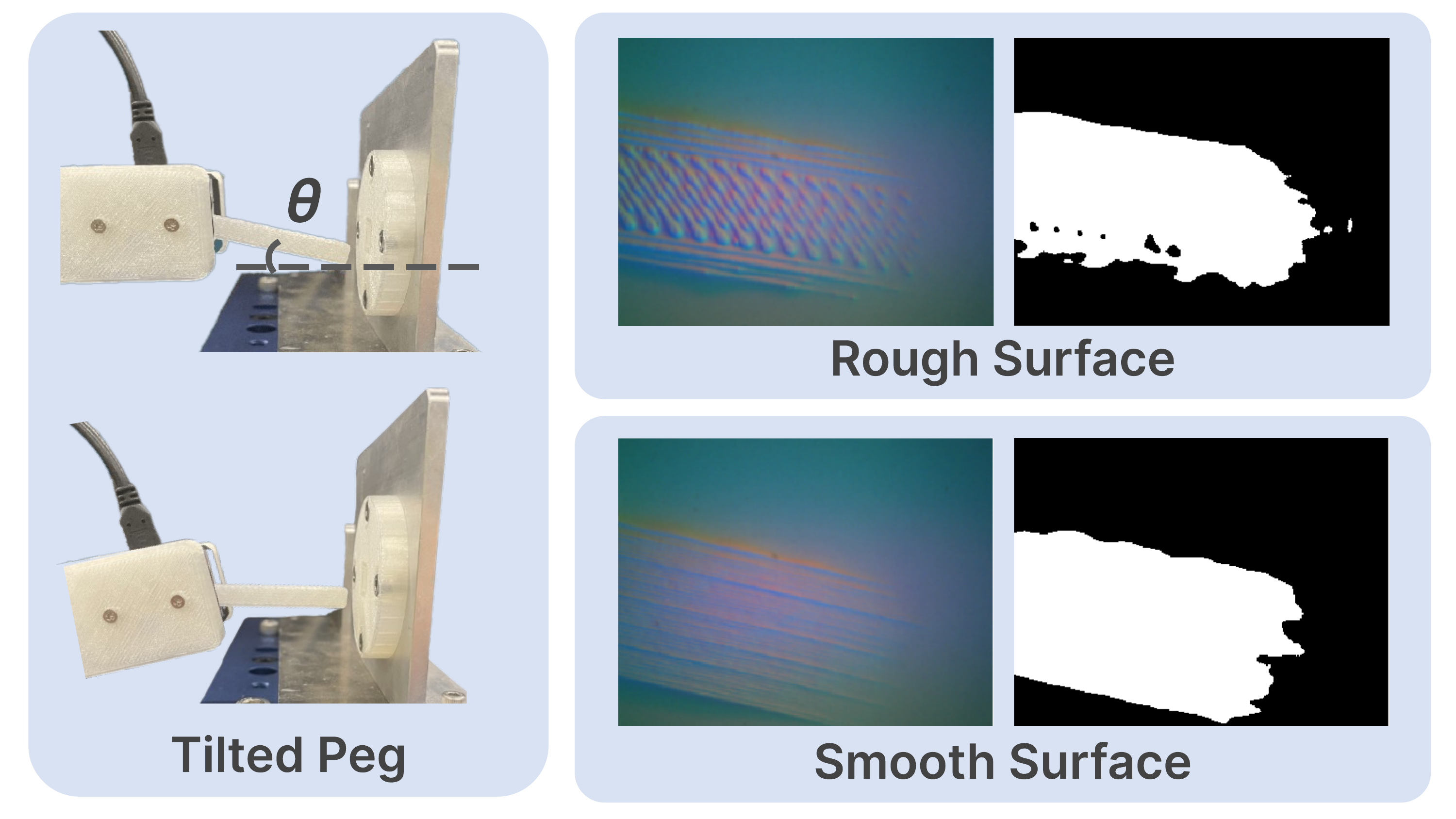}
    \caption{\textbf{Left}: Definition of $\theta$. While $\mu$ does not change after the action, $\theta$ becomes close to zero. \textbf{Right}: Depending on the surface of the peg, the estimated contact area can become noisy (top right figure), which poses challenges for accurate tilt estimation. Our proposed deep active inference approach addresses this issue by updating the internal state $\mu$ in a way that minimizes the prediction error. }
    \label{fig:roughsmooth}
\end{figure}

We model perception as the inference of an unobservable internal state $\mu$, which represents the tilt of the peg relative to the end effector. This internal state is modeled as a one-dimensional scalar. The robot's observations, denoted as $\bm{o}$, comprise a preprocessed tactile image $\bm{o}_{tac}$ and the relative angle $\theta$ between the peg and the hole. Since $\mu$ is belief of the tilt angle between the peg and the end effector, the angle $\theta$ may differ from $\mu$ in situations where the peg undergoes multiple tilts. Fig. \ref{fig:roughsmooth} illustrates the definition of $\mu$ and $\theta$. The generative model $\bm{g}$ maps this internal state to predicted tactile sensations as follows:
\begin{equation}
    \bm{o}_{tac} = \bm{g}(\mu) + \bm{\epsilon_o},
\end{equation}
\noindent where $\bm{\epsilon_o}$ is zero-mean Gaussian noise with variance $\bm{\Sigma_o}$. We assume a Gaussian prior for $\mu$ with mean zero and variance $\sigma_\mu^2$. Using these assumptions, the free energy $F$ can be expressed as:
\begin{align}
    F &= -\ln p(\bm{o}_{tac} | \mu) - \ln p(\mu) - \ln p(\theta) + \text{Const.} \label{eq:finalf}
\end{align}
\noindent Here, $p(\theta)$ is also assumed Gaussian with zero mean and variance $\sigma_\theta^2$. Under the Laplace approximation, we assume a Gaussian form for the recognition density $q(\mu)$, defined by its mean $\mu$ and variance $\sigma^2$.

Given that the agent's action of aligning its pose does not affect the internal state $\mu$, the optimal state that minimizes free energy can be found through gradient descent. The update rule is:
\begin{align}
    \dfrac{d}{dt} \mu &= - \dfrac{\partial F(\mu)}{\partial \mu} \notag\\
    &= \dfrac{\partial \bm{g}(\mu)}{\partial \mu}^T \bm{\Sigma}_{tac}^{-1} (\bm{o}_{tac} - \bm{g}(\mu)) - \sigma_\mu^{-2}\mu. \label{eq:updatemu}
\end{align}
\noindent In this equation, the term $\bm{o}_{tac} - \bm{g}(\mu)$ represents the prediction error, while $\bm{\Sigma}_{tac}^{-1}$ serves as a precision parameter. 
The expression $\partial \bm{g}(\mu)^T / \partial \mu$ maps the tactile sensory space to the internal state space, which can be obtained through a backward pass of the network when the generative model is approximated by a deep neural network \cite{PixelAI20}. 
The second term acts as a regularization term that mitigates excessive updates.
Upon updating the internal state to $\mu' = \mu + \Delta_t \dot{\mu}$, the agent performs inverse kinematics to correct the tilt by rotating the end effector around the Tool Center Point (TCP), which is defined at the center of the two fingers. 

Updating $\mu$ according to Equation (\ref{eq:updatemu}) constitutes the perceptual component of our model, as it aims to minimize the first and second terms of the free energy in Equation (\ref{eq:finalf}). Conversely, the action performed by the agent targets the minimization of the third term in the free energy equation, reflecting its goal to align the peg with the hole.
\subsubsection{\textbf{Generative Model}} 
For the generative model $\bm{g}$ (i.e. generative function) to predict tactile sensation from the inferred tilt, we approximate it using a neural network architecture closely following the design proposed in \cite{PixelAI20}, which is based on \cite{generative17}. 
The network consists of two fully connected layers followed by alternating transposed convolution and standard convolution layers, with a final 1D-Dropout layer to mitigate overfitting.
For a comprehensive understanding, readers are referred to \cite{PixelAI20}.
Our modification to the architecture involves reducing the input dimension from 4 to 1 in order to decode the tactile image from a single numerical value, which is the tilt.

\RestyleAlgo{ruled}
\begin{algorithm}[tbp]
\caption{$\bm{\pi_{align}}$ by Deep Active Inference}
\label{alg:tactileDAIF}
\textbf{\textit{Instant Decoder Training}}\\
$\bm{o}_{\text{init}} \gets \text{ContactArea}(\text{tactile image in a straight pose})$ \\
$\text{Dataset} \gets \text{DataAugmentation}(\bm{o}_{\text{init}})$ \\
$Decoder.\text{train}(\text{Dataset})$ \\
\hrulefill \\
\textbf{\textit{Active Inference Loop}} \\
$\mu \gets 0$ \\

\While{\textit{True}}{
    $\bm{o}_{\text{tac}} \gets \text{ContactArea}(\text{tactile image})$ \\
    $\bm{g}(\mu) \gets Decoder.\text{forward}(\mu)$ \\
    $\partial \bm{g} \gets Decoder.\text{backward}(\mu)$ \\
    $\bm{e}_{\text{tac}} = \bm{o}_{\text{tac}} - \bm{g}(\mu)$ \Comment{Prediction error} \\
    $\mu = \text{Update}(\mu, \partial \bm{g}, \bm{e}_{\text{tac}})$ \Comment{Perception}\\
    \If{$\mu > \text{threshold}$}{
        $\text{InverseKinematics}(\mu)$ \Comment{Action}\\
    }
}
\end{algorithm}
\subsubsection{\textbf{Instant Decoder Training}}
Our framework trains a decoder between the initial grasp and the insertion stage in just a few seconds, removing the need for a pre-trained model. Two key components enable this: contact area estimation and self-data augmentation.

We use the neural network architecture from \cite{canfnet23} to estimate the peg's contact area for each frame, as only shape information is essential for tilt inference. This serves as the first component.

The second component is self-data augmentation. A notable limitation of using vision-based tactile sensors is their susceptibility to damage. To avoid damaging the sensor surface by collecting large datasets in a real-world setting, we developed a self-data augmentation method that creates a dataset from a single tactile image of the peg in a straight pose. Assuming that we have access to this tactile image of a peg in a straight pose (aligned with the hole).

To create a dataset for decoder training, we use computer vision techniques to rotate this estimated contact area of the peg in a straight pose $\bm{o}_{init}$ by a specified degree to generate new tactile data. 
As discussed in \cite{tactileRL21}, raw tactile RGB images can be complex and contain task-irrelevant features (see Fig. \ref{fig:roughsmooth}).
Our method overcomes this issue by performing active inference based on prediction error minimization in conjunction with contact area estimation.
Extracting the contact area of the grasped peg enables the agent to quickly learn the decoder every time it grasps a new object.

The core advantage of this rapid training lies in its efficiency: our agent updates its internal state, denoted as $\mu$, by minimizing prediction error. This means the decoder does not need to capture detailed shapes of the contact area for effective performance, as illustrated in Fig. \ref{fig:architecture}. Consequently, this allows our system to swiftly adapt to new pegs without the need for a large, diverse pre-trained model.

\subsubsection{\textbf{Algorithm}}
Algorithm \ref{alg:tactileDAIF} summarizes the flow of the proposed alignment policy. 
The agent employs self-data augmentation to create a dataset. This dataset consists of tilt-labelled, estimated contact areas derived from tactile images of a peg in a straight pose $\bm{o}_{init}$. 
The decoder is trained subsequently using this dataset. During the force-controlled insertion process, a perceptual inference loop runs in real-time, updating $\mu$ following Equation (\ref{eq:updatemu}) at regular intervals. The agent takes an action only when $\mu$ exceeds a specified threshold, indicating that the peg is tilted. The action is designed to reduce the relative angle between the peg and the hole, $\theta$ becomes zero by rotating the end effector around TCP, as calculated through inverse kinematics.

\section{Experiments and Results}
\begin{figure}[tbp]
    \centering
    \includegraphics[width=\linewidth]{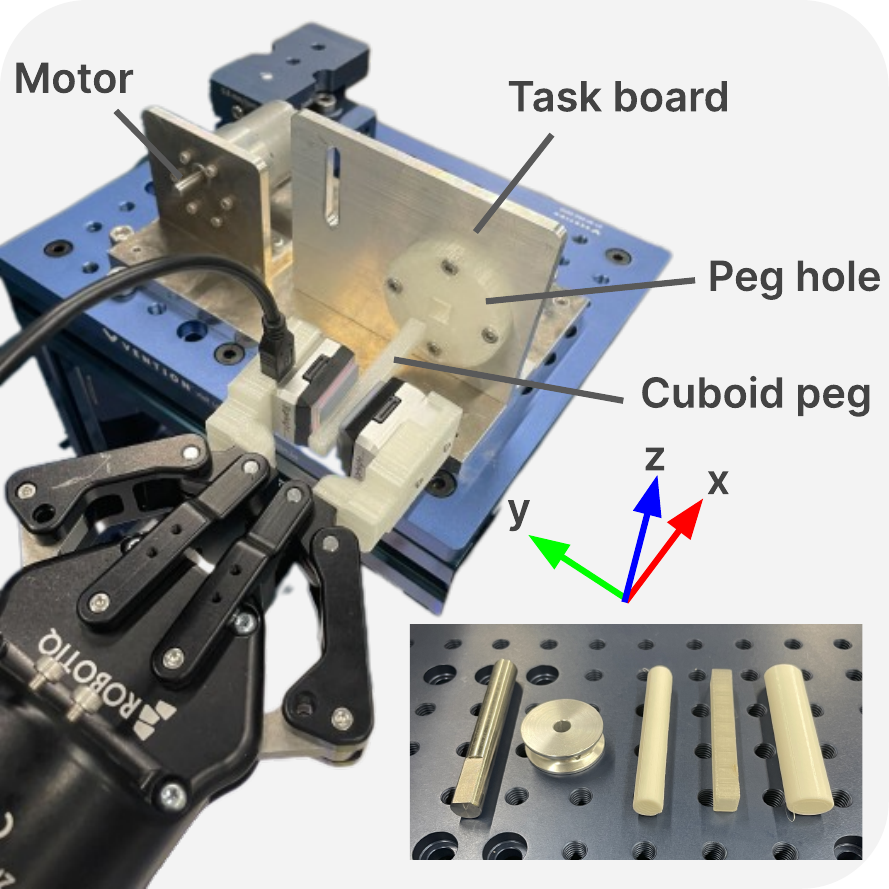}
    \caption{Overview of our experimental setup. In the bottom right corner, the pegs used in the experiments are displayed, arranged from left to right as follows: shaft, pulley, cylinder, cuboid, and elliptical cylinder. For the dual-policy experiments, we utilized two peg-in-hole tasks: inserting the pulley into the motor and inserting the cuboid into the peg hole.}
    \label{fig:experimentalsetup}
\end{figure}
\subsection{Experimental Setup}
We evaluate the proposed method on several peg-in-hole insertion tasks. We use a 6 DoF UR5 e-series robot arm with a control frequency of 500 Hz. The robotic arm has a built-in force/torque sensor at its wrist and a Robotiq 2F-85 Adaptive Gripper. Two GelSight Mini sensors are mounted on the gripper, where only one of them is used during the experiments. We use the assembly task (motor, pulley,  and shaft) from the Assembly Challenge of the World Robot Summit 2020 edition \cite{WRS2020}. Additionally, we use other 3D-printed pegs such as a cuboid, a cylinder, and an elliptical cylinder. 
For the peg-in-hole insertion tasks, we use a motor pulley and a cuboid. 
We observed that working with the shaft resulted in significant slippage along the x-axis. 
Since translations along the x-axis are not the focus of our study, we opted for a 3D-printed cuboid peg instead of the shaft.
This cuboid has an industrial-level clearance of 0.08 mm (measured with a caliper), compared to the shaft's 0.05 mm.
The motor pulley has a clearance of 0.3 mm.
The average width of the pegs is 8mm. We used Gazebo simulator \cite{gazebo04} version 9 for the simulation environment to train the insertion policy. To control the robot, the Robot Operating System (ROS) \cite{ros09} with the Universal Robot ROS Driver is used.

\subsection{Experimental Procedure}
To validate our approach, we conduct two independent experiments. 
The first one is the perceptual inference experiment, where we focus on evaluating our alignment policy's capability of inferring the tilt of multiple objects.
The second one is the active inference experiment, where the whole dual-policy system is evaluated.
\subsubsection{\textbf{Perceptual Inference Procedure}}
For this experiment, we aim to validate our alignment policy's ability to infer the tilts of multiple objects accurately.
Assuming the availability of a tactile image from a straightly grasped peg, the agent's goal is to infer peg tilts based on tactile images captured by the GelSight sensor.
The initial tactile image is processed through a pre-trained neural network model to estimate the contact area \cite{canfnet23}. We used the trained model made publicly available by the authors of \cite{canfnet23} for all tactile images in the experiments.
A training dataset (\( \mathcal{D}_{train} \)) of 500 tactile images, along with their corresponding tilts, is then created by rotating the initial contact area observation by a randomly sampled value from [-20, 20] degrees.
Test sets (\( \mathcal{D}_{test} \)) containing 100 samples are created using the tactile image in a straight pose at a different time step to avoid using the same exact shape among the train and the test data.
The decoder model was trained over five epochs, taking approximately one second on average using an Intel Core i9-9900K CPU and an Nvidia RTX-2080 Ti GPU.

Upon completing the instant decoder training, perceptual inference is performed for each of the 100 test data $\mathcal{D}_{test}$. The initial value of $\mu$ is set to 0.0, with a maximum of 500 inference iterations. The update for $\mu$ is based on Equation (\ref{eq:updatemu}) and involves two parameters: $\bm{\Sigma}^{-1}_{tac}$ and $\sigma_\mu^{^2}$. 
We empirically set \( \bm{\Sigma}^{-1}_{tac} = 2 \times 10^4 \) and \( \sigma_\mu^{^2} = 1 \times 10^{-2} \).

\subsubsection{\textbf{Dual-policy Procedure}}
Before deploying the dual-policy architecture, the force-controlled insertion policy was trained in Gazebo simulation environment. 
During the training, the agent's task was to insert a cuboid peg fixed to the end effector into a hole. The goal position, object-surface stiffness, goal pose uncertainty, and desired insertion force were randomized for each episode. The agent was trained for 100000 time steps, which takes about an hour to complete.
The active inference experiment follows the same procedure as perceptual inference for decoder training, except that we do not create test data $\mathcal{D}_{test}$. The robot initially trains the decoder using a tactile image of the peg in a straight pose. The agent creates a dataset of 1000 tactile images whose tilts were drawn randomly from [-10, 10]. We narrowed the range of degrees since we noticed that in real situations the tilts were mostly within this range.
The robot then moves to an initial position, which is about 1 cm far in $x$ axes from the hole, and [-1, 1] cm in y and z axes. This initial position is changed after 10 insertions, a total of 40 insertions were done. At this initial position, the robot grasps the peg with a random tilt, assuming that a tilt has occurred due to the contact with the environment, or initial grasp pose uncertainty. 
While grasping the peg, the perceptual inference runs in real-time with 0.5 Hz, starting from $\mu = 0$ to update until it reaches the maximum of 1000 iterations. The updated $\mu$ is used as the current belief of the internal state.
The agent first performs the active inference-based policy to align the tilted peg with the hole and transitions to the insertion policy. When $\theta$ becomes greater than a threshold, the agent switches to the alignment policy to make the peg straight. This threshold is empirically set to 0.7.

\subsection{Comparison}
To validate the adaptability and accuracy of our alignment policy on multiple objects, we also tested a baseline method that employs supervised learning with Convolutional Neural Networks (CNNs) \cite{alexnet12}. 
This baseline model is designed to predict the tilt of the peg based on the tactile image of the estimated contact area. The baseline was trained using an identical dataset and the same number of epochs as our proposed method.

For the dual-policy experiments, we employed our insertion policy without tactile feedback \cite{cristian20} as a baseline for comparison. To make fair comparison under same friction, we conducted the baseline experiments with the tactile sensor attached to the finger, although not used.

\subsection{Results}
\subsubsection{Perceptual Inference Result}

\begin{table}[tbp]
  \caption{Mean absolute error of estimated tilts [deg]}
  \label{table:result1}
  \centering
  \begin{tabular}{ccc}
    \hline
    \textbf{Object}   & \textbf{Supervised} & \textbf{Proposed method} \\
    \hline \hline
    Shaft  &  4.5  & \textbf{1.0} \\
    \hline
    Pulley & \textbf{0.3} & 1.3 \\
    \hline
    Cylinder & 2.7 & \textbf{1.5} \\
    \hline
    Cuboid & 4.4 & \textbf{1.5} \\
    \hline
    Elliptical Cylinder & \textbf{1.0} & 1.2 \\
    \hline
  \end{tabular}
\end{table}

The results of the perceptual inference experiments are summarized in Table \ref{table:result1}.
The maximum mean absolute error for estimated tilts using the supervised learning method is $4.5$ degrees, while maximum mean absolute error for our proposed method is $1.5$ degrees.
As evidenced by the results for the shaft ($4.5$ deg) and the cylinder ($2.7$ deg), the supervised learning method struggles to extract meaningful features from small datasets, especially when the peg is round and has an indistinct border in the tactile image of the contact area.
In contrast, our proposed method is robust against such noise, as the tilt inference is based on maximizing the pixel overlap between the predicted (i.e., inferred) contact area $\bm{g}(\mu)$ and the contact area derived from the actual tactile sensation.
This result indicates that our agent's inference is generally applicable to multiple pegs with different geometry.
It's worth noting that the baseline supervised learning method was only feasible because the internal state was defined as a single scalar.
In contrast, our proposed method can be applied to internal states with higher dimensions simply by altering the decoder input dimension.

\subsubsection{Dual-policy Result}
\begin{table}[tbp]
  \caption{Success rate of insertions}
  \label{table:result2}
  \centering
  \begin{tabular}{ccc}
    \hline
    \textbf{Object}   & \textbf{Without tactile feedback} & \textbf{Proposed method} \\
    \hline \hline
    Cuboid  &  1/20 (5\%)  & \textbf{36/40 (90\%)} \\
    Pulley  &  14/20 (70\%) & \textbf{37/40 (93\%)} \\
    \hline
  \end{tabular}
\end{table}

The success rates of insertions for different objects are summarized in Table. \ref{table:result2}. For the cuboid peg with a clearance of 0.08 mm, the baseline method without tactile feedback achieved a success rate of just 5\%, corresponding to 1 out of 20 trials. Given the stringent clearance, the baseline method generally failed to insert the peg with an initial random grasp pose, except in one trial where it managed to grasp the peg in almost a straight pose by chance.
In contrast, our proposed method significantly improved the success rate to 90\% with the aid of the tactile alignment policy, succeeding in 36 out of 40 trials.
Similarly, for the pulley, the baseline method achieved a relatively better but still suboptimal success rate of 70\%, or 14 out of 20 trials. 
Our proposed method again outperformed the baseline, with a success rate of 93\% based on 37 successful insertions out of 40 trials.
The relatively higher success rate of motor pulley insertions by baseline method compared with the cuboid insertions comes not only from the higher clearance (0.3 mm) but also from the shape and the material of pulley. As illustrated in bottom right of Fig. \ref{fig:experimentalsetup}, the pulley has a hole in the center, whose position itself does not change due to slippage, or rotation. We also noticed that the motor pulley insertion has less friction than 3D printed cuboid and the hole, guiding the the hole of the pulley to the motor. However, while this small friction enabled the baseline method to insert with 70\% success rate, it greatly damaged the gel of the tactile sensors. Due to this significant damage on the tactile sensor observed during the experiments, we decided to run only 20 insertions for the baseline method.

\section{Conclusion}
In this paper, we introduced a novel, tactile-based active inference approach for force-controlled peg-in-hole insertions. Our method achieved a 90\% success rate in physical robot experiments, with a clearance of less than 0.1 mm by utilizing active inference on tactile data for peg alignment. Furthermore, we demonstrated that our approach generalizes to five different objects without the need for a pre-trained model or the collection of real-world data. This addresses a significant limitation associated with the fragility of vision-based tactile sensors. As a pathway for future work, we aim to integrate vision into the system to enhance its robustness by accounting for slippage in both the insertion and rotational directions. Additionally, we see potential in expanding the internal states of our active inference model to handle more complex tasks in the tactile space.

\newpage

\bibliographystyle{unsrt}
\bibliography{main}

\begin{thebibliography}{10}

\bibitem{pegonhole}
H.~Bruyninckx, S.~Dutre, and J.~De~Schutter.
\newblock Peg-on-hole: a model based solution to peg and hole alignment.
\newblock In {\em Proceedings of 1995 IEEE International Conference on Robotics and Automation}, volume~2, pages 1919--1924 vol.2, 1995.

\bibitem{assemstrateg89}
M.E. Caine, T.~Lozano-Perez, and W.P. Seering.
\newblock Assembly strategies for chamferless parts.
\newblock In {\em Proceedings, 1989 International Conference on Robotics and Automation}, pages 472--477 vol.1, 1989.

\bibitem{review_rl_contact23}
Íñigo Elguea-Aguinaco, Antonio Serrano-Muñoz, Dimitrios Chrysostomou, Ibai Inziarte-Hidalgo, Simon Bøgh, and Nestor Arana-Arexolaleiba.
\newblock A review on reinforcement learning for contact-rich robotic manipulation tasks.
\newblock {\em Robotics and Computer-Integrated Manufacturing}, 81:102517, 2023.

\bibitem{learning_forcecontrol20}
Cristian~Camilo Beltran-Hernandez, Damien Petit, Ixchel~Georgina Ramirez-Alpizar, Takayuki Nishi, Shinichi Kikuchi, Takamitsu Matsubara, and Kensuke Harada.
\newblock Learning force control for contact-rich manipulation tasks with rigid position-controlled robots.
\newblock {\em {IEEE} Robotics and Automation Letters}, 5(4):5709--5716, oct 2020.

\bibitem{gelsight17}
Wenzhen Yuan, Siyuan Dong, and Edward~H. Adelson.
\newblock Gelsight: High-resolution robot tactile sensors for estimating geometry and force.
\newblock {\em Sensors (Basel, Switzerland)}, 2017.

\bibitem{cristian20}
Cristian~C. Beltran-Hernandez, Damien Petit, Ixchel~G. Ramirez-Alpizar, and Kensuke Harada.
\newblock Variable compliance control for robotic peg-in-hole assembly: A deep-reinforcement-learning approach.
\newblock {\em Applied Sciences}, 10(19), 2020.

\bibitem{gelslim18}
Elliott Donlon, Siyuan Dong, Melody Liu, Jianhua Li, Edward Adelson, and Alberto Rodriguez.
\newblock Gelslim: A high-resolution, compact, robust, and calibrated tactile-sensing finger.
\newblock In {\em 2018 IEEE/RSJ International Conference on Intelligent Robots and Systems (IROS)}, pages 1927--1934, 2018.

\bibitem{digit20}
Mike Lambeta, Po-Wei Chou, Stephen Tian, Brian Yang, Benjamin Maloon, Victoria~Rose Most, Dave Stroud, Raymond Santos, Ahmad Byagowi, Gregg Kammerer, Dinesh Jayaraman, and Roberto Calandra.
\newblock Digit: A novel design for a low-cost compact high-resolution tactile sensor with application to in-hand manipulation.
\newblock {\em IEEE Robotics and Automation Letters}, 2020.

\bibitem{omnitact20}
Akhil Padmanabha, Frederik Ebert, Stephen Tian, Roberto Calandra, Chelsea Finn, and Sergey Levine.
\newblock Omnitact: A multi-directional high-resolution touch sensor.
\newblock {\em 2020 IEEE International Conference on Robotics and Automation (ICRA)}, pages 618--624, 2020.

\bibitem{Bauz2019TactileMA}
Maria Bauz{\'a}, Oleguer Canal, and Alberto Rodriguez.
\newblock Tactile mapping and localization from high-resolution tactile imprints.
\newblock {\em 2019 International Conference on Robotics and Automation (ICRA)}, pages 3811--3817, 2019.

\bibitem{Bauz2corl}
Maria Bauz{\'a}, Eric Valls, Bryan Lim, Theo Sechopoulos, and Alberto Rodriguez.
\newblock Tactile object pose estimation from the first touch with geometric contact rendering.
\newblock In {\em Conference on Robot Learning}, 2020.

\bibitem{poseaamas22}
Tarik Kelestemur, Robert Platt, and Taskin Padir.
\newblock Tactile pose estimation and policy learning for unknown object manipulation.
\newblock In Piotr Faliszewski, Viviana Mascardi, Catherine Pelachaud, and Matthew~E. Taylor, editors, {\em 21st International Conference on Autonomous Agents and Multiagent Systems, {AAMAS} 2022, Auckland, New Zealand, May 9-13, 2022}, pages 742--750. International Foundation for Autonomous Agents and Multiagent Systems {(IFAAMAS)}, 2022.

\bibitem{dense19}
Siyuan Dong and Alberto Rodriguez.
\newblock Tactile-based insertion for dense box-packing.
\newblock In {\em 2019 IEEE/RSJ International Conference on Intelligent Robots and Systems (IROS)}, pages 7953--7960, 2019.

\bibitem{safe23}
Letian Fu, Huang Huang, Lars Berscheid, Hui Li, Ken Goldberg, and Sachin Chitta.
\newblock Safe self-supervised learning in real of visuo-tactile feedback policies for industrial insertion.
\newblock In {\em 2023 IEEE International Conference on Robotics and Automation (ICRA)}, pages 10380--10386, 2023.

\bibitem{tactileRL21}
Siyuan Dong, Devesh Jha, Diego Romeres, Sangwoon Kim, Daniel Nikovski, and Alberto Rodriguez.
\newblock Tactile-rl for insertion: Generalization to objects of unknown geometry.
\newblock In {\em 2021 IEEE International Conference on Robotics and Automation (ICRA)}, 2021.

\bibitem{canfnet23}
Niklas Funk, Paul-Otto Müller, Boris Belousov, Anton Savchenko, Rolf Findeisen, and Jan Peters.
\newblock Canfnet: High-resolution pixelwise contact area and normal force estimation for visuotactile sensors using neural networks.
\newblock 2023.

\bibitem{friston10}
Karl Friston.
\newblock Friston, k.j.: The free-energy principle: a unified brain theory? nat. rev. neurosci. 11, 127-138.
\newblock {\em Nature reviews. Neuroscience}, 11:127--38, 02 2010.

\bibitem{survey_lanillos2021}
Pablo Lanillos, Cristian Meo, Corrado Pezzato, Ajith~Anil Meera, Mohamed Baioumy, Wataru Ohata, Alexander Tschantz, Beren Millidge, Martijn Wisse, Christopher~L. Buckley, and Jun Tani.
\newblock Active inference in robotics and artificial agents: Survey and challenges, 2021.

\bibitem{PixelAI2020}
Cansu Sancaktar, Marcel A.~J. van Gerven, and Pablo Lanillos.
\newblock End-to-end pixel-based deep active inference for body perception and action.
\newblock In {\em 2020 Joint {IEEE} 10th International Conference on Development and Learning and Epigenetic Robotics ({ICDL}-{EpiRob})}, 2020.

\bibitem{deepaif18}
Kai Ueltzhöffer.
\newblock Deep active inference.
\newblock {\em Biological Cybernetics}, 112(6):547--573, oct 2018.

\bibitem{deepaif_MC20}
Zafeirios Fountas, Noor Sajid, Pedro Mediano, and Karl Friston.
\newblock Deep active inference agents using monte-carlo methods.
\newblock In H.~Larochelle, M.~Ranzato, R.~Hadsell, M.~F. Balcan, and H.~Lin, editors, {\em Advances in Neural Information Processing Systems}, volume~33, pages 11662--11675. Curran Associates, Inc., 2020.

\bibitem{deepaif_POMDP20}
Otto van~der Himst and Pablo Lanillos.
\newblock Deep active inference for partially observable mdps.
\newblock In {\em International Workshop on Affective Interactions}, 2020.

\bibitem{deepaif_vpg20}
Beren Millidge.
\newblock Deep active inference as variational policy gradients.
\newblock {\em Journal of Mathematical Psychology}, 96:102348, 2020.

\bibitem{adaptivebody18}
Pablo Lanillos and Gordon Cheng.
\newblock Adaptive robot body learning and estimation through predictive coding.
\newblock In {\em 2018 {IEEE}/{RSJ} International Conference on Intelligent Robots and Systems ({IROS})}. {IEEE}, oct 2018.

\bibitem{PixelAI20}
Cansu Sancaktar, Marcel A.~J. van Gerven, and Pablo Lanillos.
\newblock End-to-end pixel-based deep active inference for body perception and action.
\newblock In {\em 2020 Joint {IEEE} 10th International Conference on Development and Learning and Epigenetic Robotics ({ICDL}-{EpiRob})}, 2020.

\bibitem{novelmanip20}
Corrado Pezzato, Riccardo Ferrari, and Carlos Hernández~Corbato.
\newblock A novel adaptive controller for robot manipulators based on active inference.
\newblock {\em IEEE Robotics and Automation Letters}, PP:1--1, 02 2020.

\bibitem{empiricalhumanoid22}
Guillermo Oliver, Pablo Lanillos, and Gordon Cheng.
\newblock An empirical study of active inference on a humanoid robot.
\newblock {\em IEEE Transactions on Cognitive and Developmental Systems}, 14(2):462--471, 2022.

\bibitem{domain17}
Josh Tobin, Rachel Fong, Alex Ray, Jonas Schneider, Wojciech Zaremba, and Pieter Abbeel.
\newblock Domain randomization for transferring deep neural networks from simulation to the real world.
\newblock In {\em 2017 IEEE/RSJ International Conference on Intelligent Robots and Systems (IROS)}, pages 23--30, 2017.

\bibitem{generative17}
Alexey Dosovitskiy, Jost Springenberg, Maxim Tatarchenko, and Thomas Brox.
\newblock Learning to generate chairs, tables and cars with convolutional networks.
\newblock {\em IEEE Transactions on Pattern Analysis and Machine Intelligence}, 39:1--1, 05 2016.

\bibitem{WRS2020}
{World Robot Summit}.
\newblock {Assembly Challenge 2020}, 2020.
\newblock Accessed: 2023-9-15.

\bibitem{gazebo04}
N.~Koenig and A.~Howard.
\newblock Design and use paradigms for gazebo, an open-source multi-robot simulator.
\newblock In {\em 2004 IEEE/RSJ International Conference on Intelligent Robots and Systems (IROS) (IEEE Cat. No.04CH37566)}, volume~3, pages 2149--2154 vol.3, 2004.

\bibitem{ros09}
Morgan Quigley, Ken Conley, Brian Gerkey, Josh Faust, Tully Foote, Jeremy Leibs, Rob Wheeler, and Andrew Ng.
\newblock Ros: an open-source robot operating system.
\newblock volume~3, 01 2009.

\bibitem{alexnet12}
Alex Krizhevsky, Ilya Sutskever, and Geoffrey~E. Hinton.
\newblock Imagenet classification with deep convolutional neural networks.
\newblock In {\em Proceedings of the 25th International Conference on Neural Information Processing Systems - Volume 1}, NIPS'12, page 1097–1105, Red Hook, NY, USA, 2012. Curran Associates Inc.

\end{thebibliography}

\end{document}